\documentclass{ieeeaccess}
\usepackage{helvet}

\usepackage[T1]{fontenc}
\usepackage{times}

\usepackage{natbib}
\usepackage{multibib} 
\usepackage{amsmath,amssymb,amsfonts}
\usepackage{algorithmic}
\usepackage{graphicx}
\usepackage{float}
\usepackage{booktabs}
\usepackage{multirow}
\usepackage{rotating}
\usepackage{colortbl}
\usepackage{url}
\usepackage{silence}
\usepackage[table]{xcolor}

\usepackage{soul}
\usepackage{mathptmx}
\usepackage{latexsym}
\usepackage{textcomp}
\usepackage{comment}
\def\BibTeX{{\rm B\kern-.05em{\sc i\kern-.025em b}\kern-.08em
    T\kern-.1667em\lower.7ex\hbox{E}\kern-.125emX}}

\newcommand{\mrs}[1]{#1}

\begin{document}
\history{Date of publication xxxx 00, 0000, date of current version xxxx 00, 0000.}
\doi{10.1109/ACCESS.2017.DOI}

\title{Measuring Catastrophic Forgetting in Cross-Lingual Classification: Transfer Paradigms and Tuning Strategies}

\author{\uppercase{Boshko Koloski}\authorrefmark{1,2}, \uppercase{Bla\v{z} \v{S}krlj}\authorrefmark{1}, \uppercase{Marko Robnik-\v{S}ikonja}\authorrefmark{3}, and \uppercase{Senja Pollak}\authorrefmark{1}}
\address[1]{Jožef Stefan Institute, Ljubljana, Slovenia (e-mail: example@ijs.si)}
\address[2]{Jožef Stefan International Postgraduate School, Ljubljana, Slovenia}
\address[3]{Faculty of Computer and Information Science, University of Ljubljana, Ljubljana, Slovenia}

\tfootnote{The authors acknowledge the financial support from the Slovenian Research Agency for research core funding (No. P2-0103 and P6-0411) and the projects: Computer-assisted multilingual news discourse analysis with contextual embeddings (CANDAS, J6-2581), Hate speech in contemporary conceptualizations of nationalism, racism, gender and migration (SOVRAG, J5-3102), Embeddings-based techniques for Media Monitoring Applications (EMMA, L2-50070) and Large Language Models for Digital Humanistics (LLM4DH, GC-0002). A Young Researcher Grant PR-12394 supported the work of the first author.}

\markboth
{Koloski \headeretal: Measuring Catastrophic Forgetting in Cross-Lingual Classification}
{Koloski \headeretal: Measuring Catastrophic Forgetting in Cross-Lingual Classification}

\corresp{Corresponding author: B. Koloski (e-mail: boshko.koloski@ijs.si).}

\begin{abstract}
\mrs{Cross-lingual transfer leverages knowledge from a resource-rich source language, commonly English, to enhance performance in less-resourced target languages. Two widely used strategies are: \textbf{Cross-Lingual Validation (CLV)}, which involves training on the source language and validating on the target language, and \textbf{Intermediate Training (IT)}, where models are first fine-tuned on the source language and then further trained on the target language. While both strategies have been studied, their effects on encoder-based models for classification tasks remain underexplored. In this paper, we systematically compare these strategies across six multilingual classification tasks, evaluating downstream performance, catastrophic forgetting, and both zero-shot and full-shot scenarios. Additionally, we contrast parameter-efficient adapter methods with full-parameter fine-tuning. Our results show that IT generally performs better in the target language, whereas CLV more effectively preserves source-language knowledge across multiple cross-lingual transfers. These findings underscore the trade-offs between optimizing target performance and mitigating catastrophic forgetting.}
\end{abstract}

%% OLD: The cross-lingual transfer is a promising technique to solve tasks in less-resourced languages. This study compares two fine-tuning approaches combined with zero-shot and full-shot learning approaches for large language models in a cross-lingual setting. As fine-tuning strategies, we compare parameter-efficient adapter methods with fine-tuning of all parameters. As cross-lingual transfer strategies, we compare the intermediate-training (IT ) that uses each language sequentially and cross-lingual validation (CLV ) that uses a target language already in the validation phase of fine-tuning. We assess the success of transfer and the extent of catastrophic forgetting in a source language due to cross-lingual transfer, i.e., how much previously acquired knowledge is lost when we learn new information in adifferent language. The results on different classification problems, each containing datasets in several languages, show that the IT cross-lingual strategy outperforms CLV for the target language. Our findings indicate that, in the majority of cases, the CLV strategy demonstrates superior retention of knowledge in the base language (English) compared to the IT strategy when evaluating catastrophic forgetting in multiple cross-lingual transfers.


\titlepgskip=-15pt

\maketitle
\section{Introduction}

Transfer learning has emerged as one of the most popular paradigms in deep learning. It aims to transfer already learned neural network weights from one task or model to another. With the emergence of pre-trained large language models \textit{(LLMs)}\footnote{Clarification: Throughout this study, we refer to LLMs as pretrained models that, by recent standards, are considered small. This choice is made because these models use the same transformer architecture and learning tasks for language modeling, differing primarily in the number of parameters.} in monolingual and multilingual settings, such as BERT \citep{devlin-etal-2019-bert}, GPT-3 \citep{brown2020language}, and XLM-R \citep{NEURIPS2019_c04c19c2}, they have taken the field of NLP by storm. LLMs are pre-trained with self-supervised learning, where the idea is to learn the data distribution without explicit labels. For example, models are asked to solve fill-a-gap tasks in natural language settings (\textit{Masked Language Modeling (MLM)}). In \textit{Causal Language Modeling (CLM)}, the model predicts the next word based on the "cause"—the input provided so far. \citet{NEURIPS2019_c04c19c2} introduced the task of \textit{Translated Language Modeling (TLM)}, where masked words are predicted in two parallel sentences in different languages, improving language alignment. Recently, \citet{pmlr-v235-zhang24m} conducted a theoretical comparison between masked language models (MLMs) and causal language models (CLMs). They found that, in classification tasks, MLMs exhibit superior performance due to their flexible token targeting, which fosters more inter-sample connections than the fixed token positions in autoregressive models. Building on these findings, we focus on MLM-based models, such as XLM-R \citep{NEURIPS2019_c04c19c2}, which provide a strong foundation across a broad range of languages.

Large Language Models (LLMs) are trained on vast amounts of data and effectively capture linguistic structures, enabling them to perform well in zero-shot and few-shot learning settings \citep{brown2020language, wei2021finetuned, kojima2022large}. For a model to specialize in a specific downstream task, only a relatively small amount of task-specific data is required. The strong generalization ability of LLMs from a low number of examples makes them a suitable approach for knowledge transfer in low-resource settings, particularly when data from high-resource languages is available. These models are predominantly pretrained on English-language datasets, making an efficient transfer from English downstream datasets highly desirable \citep{chen2021model}. \mrs{When knowledge is transferred for a specific task or set of tasks from one language to another language, this process is referred to as \textbf{cross-lingual transfer}.}

A common problem in transfer learning, where knowledge is transferred to another problem, is \textbf{catastrophic forgetting} (CF) \citep{mccloskey1989catastrophic, kemker2018measuring}, where models forget previously acquired knowledge when adapted to a novel task.

We differentiate between three cross-lingual transfer approaches: a zero-shot transfer and two full-shot strategies. In \textbf{zero-shot transfer}, we assume that the model has already acquired task-specific knowledge during training in the source language, and we directly employ the trained model on the same task in the target language \citep{pelicon-etal-2021-zero,wang2019cross,wang2019survey,koloski2022out,winata-etal-2022-cross}. \textbf{Intermediate training} is a full-shot strategy, where the model is first trained in the source language, followed by fine-tuning on the target language \citep{zhao2020closer, pelicon2021investigating}. \textbf{Cross-lingual validation}, also a full-shot strategy, involves training the model on the source language data while using the target language data as the validation set. In our experiments, we evaluate the performance of all methods on unseen test data in the \texttt{target} language. \textbf{Figure~\ref{fig1} showcases the two cross-lingual paradigms.}

In reusing LLMs for different tasks, \textbf{adapters} \citep{NIPS2017_e7b24b11, pmlr-v97-houlsby19a} were introduced to avoid updating all of the pretrained model's weights when fine-tuning to a new problem. The idea is to fine-tune only a specific section of the model in a parameter-efficient manner.

In this work, we investigated the following research questions:
\begin{itemize}
    \item How do two different cross-lingual training paradigms, intermediate training and cross-lingual validation, influence cross-lingual transfer results?
    \item Is full-model fine-tuning better compared to adapters when it comes to \textit{cross-lingual learning} and \textit{catastrophic forgetting}?
    \item How does \textit{catastrophic forgetting} affect previously acquired knowledge in multiple transfer episodes?
    \item In a low-resource (compute-wise) setting, which cross-lingual training paradigm yields better results: intermediate training or cross-lingual validation?
\end{itemize}

Our contributions are as follows: 
\begin{enumerate}
    \item We present the first study examining the effect of catastrophic forgetting on different cross-lingual paradigms.
    \item We systematically evaluate two different cross-lingual training regimes: intermediate training and cross-lingual validation.
    \item We measure the effect of catastrophic forgetting with well-established metrics and provide a blueprint for selecting a metric for cross-lingual training when retaining performance on the source language is essential.
    \item We prepare open-source cross-lingual adapters for multiple tasks in three less-resourced languages.
\end{enumerate}

We describe the related work in Section \ref{sec:rel_work}, followed by the description of the cross-lingual methodology in Section \ref{sec:meth}, and the experimental setup in Section \ref{sec:exp}. We summarize the results in Section \ref{sec:res} and present conclusions in Section \ref{sec:final}.

\Figure[t!](topskip=0pt, botskip=0pt, midskip=0pt)[width=0.9\linewidth] {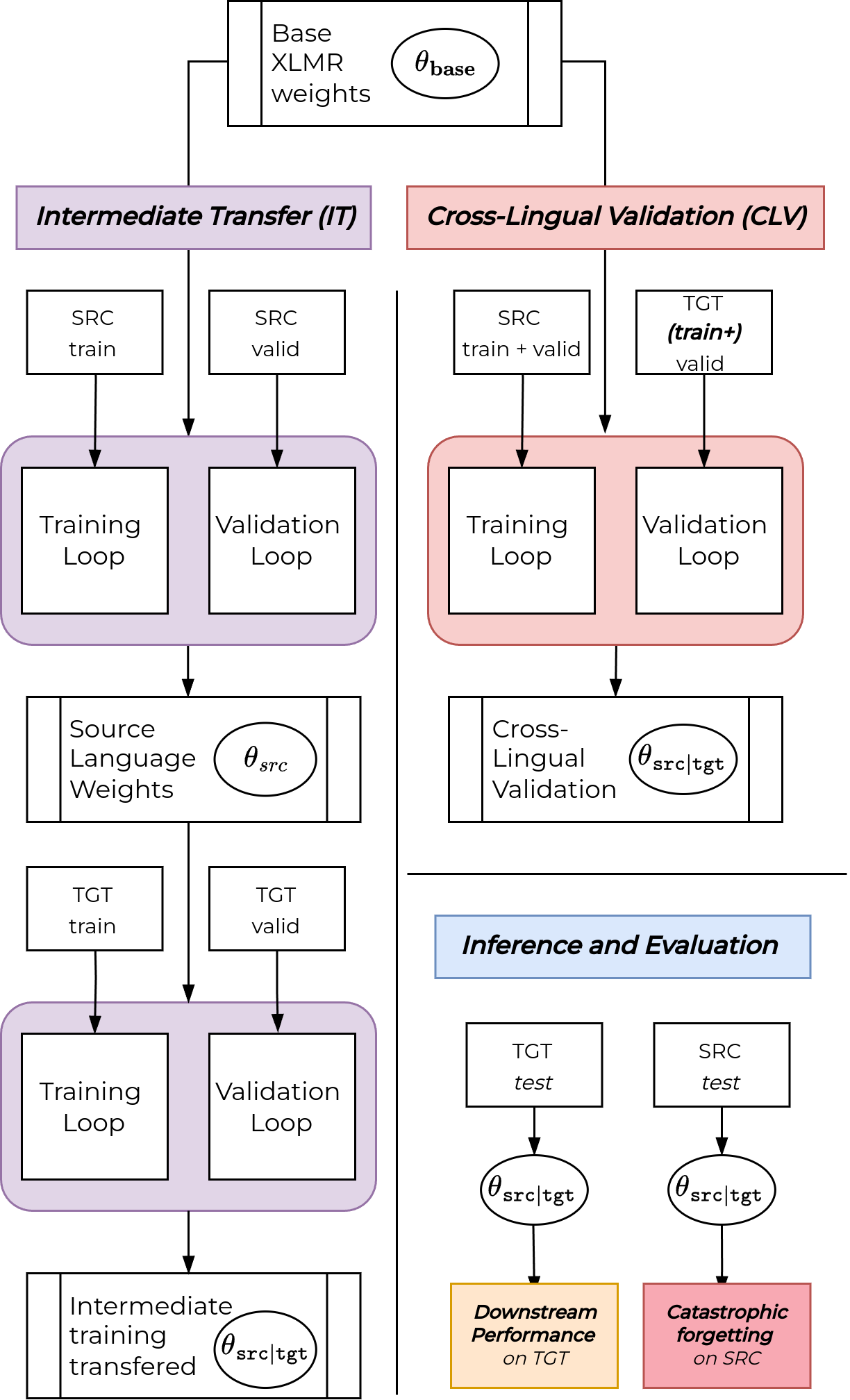} {\mrs{Comparison of two cross-lingual training approaches. Both methods begin by aligning the base weights to the source language. The cross-lingual validation (CLV) approach updates weights while validating on the target language. The intermediate-training (IT) approach initially trains and validates on the source language before transitioning to the target language. The zero-shot approach uses the source-adapted weights to infer on the test dataset. Both approaches yield target-language-adapted weights. Downstream performance is evaluated on the target dataset, while catastrophic forgetting is measured on the source dataset.} \label{fig1}}

\section{Related work}
\label{sec:rel_work}
In this section, we discuss related work, split into three parts: cross-lingual transfer, adapters, and catastrophic forgetting.

\subsection{Cross-lingual transfer}
The pioneering work on cross-lingual modeling focused on aligning static word embeddings between languages to force words with the same meaning to be as close as possible in the vector space. \citet{Mikolov2013ExploitingSA} aligned models with a linear transformation in the Word2Vec \citep{NIPS2013_9aa42b31} embedding space between languages. \citet{DBLP:conf/iclr/LampleCRDJ18} focused on utilizing GAN model to train linear mapping between static vector spaces.  \citet{ulvcar2022cross} constructed non-linear mapping between contextual ELMo embeddings  \citep{peters-etal-2018-deep} suing GANs. \citet{NEURIPS2019_c04c19c2} introduced the XLM-R model trained with the translated language modeling objective to better suit cross-lingual transfer.
\citet{van-der-heijden-etal-2021-multilingual} formulated the cross-lingual classification as a meta-learning adaptation to new tasks \citep{schmidhuber1987evolutionary} where they treat each language as a different task, showcasing promising results in the limited-resource scenario. \citet{wang-etal-2021-cross-lingual} treated cross-lingual classification as a node-classification task. They constructed a multi-layer graph based on document similarity, word co-occurrence, and lexical structure and initialized the nodes with XLM \citep{NEURIPS2019_c04c19c2}. They then applied a convolutional graph neural network  \citep{kipf2017semi} to the graph structure and reported improved performance compared to the base model. \citet{zhao2020closer} study the selection of instances in a few-shot scenario and show that methods are sensitive to the quality of the annotated data. Recently, \citet{cooper-stickland-etal-2023-robustification} proposed an effective pretraining strategy based on modeling typological, grammatical, or morphological noise in the data that boosts the cross-lingual zero-shot performance.

 \subsection{Adapters} 
To avoid updating and storing all the parameters of an LLM during fine-tuning to a new task, adapters \citep{NIPS2017_e7b24b11, pmlr-v97-houlsby19a} fine-tune only a specific section of the model in a parameter-efficient manner. Adapters have demonstrated encouraging results in adapting to new tasks  \citep{pmlr-v97-houlsby19a}, domains \citep{bapna2019simple}, and languages \citep{pfeiffer2020mad, ansell2021mad} while being highly efficient and lightweight \citep{ruckle2020adapterdrop}.

%new part:
\mrs{\citet{hu2022lora} introduced Low-Rank Adaptation (LoRA) as a more efficient method for fine-tuning Large Language Models (LLMs) on new tasks. LoRA decomposes model weights into multiple low-rank matrices, trains them separately, and then recombines them before adding them back to the original pre-trained weights. Building on this idea, \citet{dettmers2023qlora} proposed quantized LoRA, which reduces the model’s footprint by projecting it into lower dimensions, and \citet{valipour-etal-2023-dylora} presented a rank-free, dynamic LoRA setup. Across all variants, LoRA-based adaptations have consistently demonstrated faster update times, reduced memory requirements, and, in some cases, improved downstream performance compared to full-parameter tuning. However, systematic evaluation on cross-lingual tasks still remains unaddressed. 
}

 \subsection{Catastrophic forgetting} 
 Catastrophic forgetting (CF) \citep{mccloskey1989catastrophic} is a general term for forgetting previously acquired knowledge in machine learning when the model is adapted to a novel task. To overcome the problem, researchers need to opt whether to optimize the model's \textit{stability} - the ability to retain acquired knowledge, or the model's \textit{plasticity} - the ability to learn new information effectively. %These approaches can be divided into ones that regularize the deviation between the initial and fine-tuned weights and ones that introduce additional signals into the model to aid in switching between problems. 
 
\citet{sun2019fine} propose choosing a lower learning rate to overcome catastrophic forgetting in neural networks. \citet{xu2020forget} explore regularization strategies by introducing selective (Elastic Weight Consolidation by
\citet{kirkpatrick2017overcoming}) and non-selective ($\lambda2$ regularisation by \citet{l2regs}) regularisation terms between the initial and the fine-tuned weights. They see a boost in performance in domain adaptation, task transfer, and continuous learning settings with both regimes. \citet{yang2020towards} introduce concerted training consisting of model distillation to retain the previous knowledge, a dynamic switching gate to avoid catastrophic forgetting of pre-trained knowledge, and a scheduled policy to adjust the learning rate. Their approach shows promising results for the machine translation task. \citet{vu2022overcoming} propose overcoming CF by prompt tuning and improve performance over classical fine-tuning when transferring between less-related languages. They opt between mixing unlabeled multilingual data in the prompt tuning or explicitly factoring prompts into composable language and task components.

\mrs{Model-wise catastrophic forgetting affects both small and large language models \citep{kotha2024understanding}, as well as multimodal models \citep{zhai2023investigating}. \citet{kotha2024understanding} showed that steps like instruction tuning and reinforcement learning improve performance on tasks seen during fine-tuning but hurt performance on others. This suggests that models learn to focus on the training tasks at the expense of their broader capabilities. \citet{ren2024analyzing} found that low-rank adaptations for downstream tasks can worsen catastrophic forgetting for LLMs, but using interpolation-based updates between source and target weights can help. However, they only studied English-to-English transfers from large decoder-based models and didn't utilise established metrics for CF assessment.  It is still unknown how parameter-efficient tuning affects catastrophic forgetting in cross-lingual scenarios, which motivates our current work.}
\section{Cross-lingual Transfer}
\label{sec:meth}

Following \citet{winata-etal-2022-cross}, we denote the representation of a language model with parameters $\theta$ and the dataset $D$ in a given language $L$ as $D_{L}$. Each $D_{L}$ consists of tuples of documents $x$ and labels $y$: 
$D_{L} = \{(x_{1},y_{1}), \cdots, (x_{i},y_{i}), \cdots, (x_{N},y_{N})\}$, where $x_i$ is the $i$-th document in a collection of $N$ documents for the task in the given language. In the cross-lingual setting, we distinguish between the source language $L_{\texttt{src}}$, used for fine-tuning the initial pretrained $\theta$, and the target language $L_{\texttt{tgt}}$, used to evaluate cross-lingual transfer with $\theta$. The data for each language is split into three parts, used in different phases of fine-tuning and evaluation:
\begin{itemize}
    \item \textbf{\texttt{train}}: the data used for training, i.e., fine-tuning the models.
    \item \textbf{\texttt{valid}}: the data used to validate the models, i.e., measuring performance during fine-tuning and selecting hyperparameters.
    \item \textbf{\texttt{test}}: the data used for testing the models. This split is used to compare models and is not utilized during training or validation.
\end{itemize}

We next describe the cross-lingual transfer approaches in detail, distinguishing between zero-shot transfer (without considering any data in the target language) and two different strategies when target language data is available.

\subsection{Zero-shot Cross-lingual Transfer}

In zero-shot (\textit{ZS}) cross-lingual transfer, a pretrained LLM is fine-tuned using data from a single language $L_{\texttt{src}}^{\texttt{train}}$ and validated on the same language $L_{\texttt{src}}^{\texttt{valid}}$ to obtain the model $\theta_{\texttt{src}}$. The zero-shot transfer performance is then measured on $L_{\texttt{tgt}}^{\texttt{test}}$.

\subsection{Intermediate-training Transfer}

In intermediate-training full-shot transfer (\textit{IT}), the model undergoes a two-phase training process. First, the model is fine-tuned on data from a resource-rich language, using $L_{\texttt{src}}^{\texttt{train}}$, and validated on $L_{\texttt{src}}^{\texttt{valid}}$ to obtain $\theta_{\texttt{src}}$. In the second phase, this model is further fine-tuned on $L_{\texttt{tgt}}^{\texttt{train}}$ and validated on $L_{\texttt{tgt}}^{\texttt{valid}}$ to obtain $\theta_{\texttt{src} \rightarrow \texttt{tgt}}$. This approach uses multiple languages \emph{sequentially}.

\subsection{Cross-lingual Validation Transfer}

In cross-lingual validation transfer (\textit{CLV}), the model $\theta_{\texttt{src}|\texttt{tgt}}$ is first fine-tuned on data from the source language and validated on the validation set from the target language. This approach incorporates the target language during fine-tuning and emphasizes training in multiple languages. Additionally, \textit{CLV} can result in faster training when the goal is to produce a single model for two languages. 

We distinguish between two \textit{CLV} approaches: \emph{valid} (a few-shot setting) and \emph{valid+train}. In the \emph{valid} approach, only $L_{\texttt{tgt}}^{\texttt{valid}}$ is used. In the \emph{valid+train} approach, the \texttt{train} and \texttt{valid} splits are merged, forming $L_{\texttt{tgt}}^{\texttt{merged}} = L_{\texttt{tgt}}^{\texttt{train}} \cup L_{\texttt{tgt}}^{\texttt{valid}}$. Similarly, for the source language, $L_{\texttt{src}}^{\texttt{merged}} = L_{\texttt{src}}^{\texttt{train}} \cup L_{\texttt{src}}^{\texttt{valid}}$ is used. For a fair comparison, the \emph{valid+train} variant of \textit{CLV} is directly comparable with \textit{IT}, as the same amount of source and target language data is available.

%In case of combined splits, we train the model on the $L_{\texttt{src}}^{\texttt{merged}}$ and use  $L_{\texttt{tgt}}^{\texttt{merged}}$ for validation. 
\section{Experimental Setting}
\label{sec:exp}

This section presents our empirical study, starting with the classification datasets and three experimental setups, followed by a description of hyperparameters. Our code will be freely available after de-anonymization.

\subsection{Data}
Our study uses six problems: four binary and two multi-label problems. Table \ref{tab:data_stats} provides detailed dataset statistics. For each dataset, we use the official train-validation-test splits unless stated otherwise. 

\subsubsection{Hate-Speech Dataset}

We follow the multilingual hate-speech dataset construction by \citet{pelicon-etal-2021-zero}. This approach combines data from various sources to build a hate-speech classification dataset using social media posts across five languages. The languages involved are English (\texttt{en}) \citep{zampieri2019predicting}, German (\texttt{ge}) \citep{wiegand2018overview}, Arabic (\texttt{ar}) \citep{zampieri2020semeval}, Slovenian (\texttt{sl}) \citep{frenk}, and Croatian (\texttt{hr}) \citep{shekhar2020automating, pollak2021embeddia}. In our cross-lingual approach, we designate English (\texttt{en}) as the source language (\textit{\texttt{src}}), and the remaining languages as the target languages (\textit{\texttt{tgt}}). We refer to this dataset as \texttt{HateSpeech}.

\subsubsection{Reviews Datasets}
The dataset of Amazon reviews comprises sentiment analysis for three review categories: \texttt{DVD}, \texttt{Books}, and \texttt{Music}. Each category contains reviews in English (\texttt{en}), Japanese (\texttt{jp}), German (\texttt{ge}), and French (\texttt{fr}). Following previous studies \citep{xu2017cross, fei2020cross, wang2021cross}, we converted the labels into binary ones (the original labels ranged from 0 to 5) by applying a threshold at \textbf{3}, mapping 0, 1, 2 to the negative class and 3, 4, 5 to the positive class. Since the original datasets only defined train and test splits, we randomly selected 80\% of the train instances for training and 20\% for validation per language and per category. We refer to the averaged statistics of these datasets as Reviews* in some explanations.

\subsubsection{SLU Dataset}
The SLU (spoken language understanding) dataset \citep{schuster-etal-2019-cross-lingual} contains brief, task-focused English (\texttt{en}), Spanish (\texttt{es}), and Thai (\texttt{th}) utterances. We assess classification on \textbf{12} distinct intent types using this dataset.

\subsubsection{XGLUE Dataset}
The XGLUE \citep{liang-etal-2020-xglue} dataset is a subset of the cross-lingual version of the GLUE benchmark. The news classification subtask aims to categorize news articles written in five languages: English (\texttt{en}), Spanish (\texttt{es}), French (\texttt{fr}), German (\texttt{ge}), and Russian (\texttt{ru}) into \textbf{10} categories. While the original dataset only specified validation and test splits for languages other than English, we followed a similar approach to the Reviews dataset, randomly allocating 80\% of the original validation instances for training and the remaining 20\% for validation.

\begin{table}[ht]
    \centering
    \resizebox{\columnwidth}{!}{\begin{tabular}{l|cccccc}
\toprule
       &   &   &  &   &   max &   min  \\
       dataset &  lang. &  instances & len. &  classes &  support &   support \\
             &   &   &  &   &  label &  label \\
\midrule
Reviews{*}   &          \texttt{en} &       4000 &  845.16 &        2 &   50.00 &    50.00 \\
&         \texttt{ge} &       4000 &  865.57 &        2 &   50.00 &    50.00 \\            
&          \texttt{fr}  &       4000 &  641.01 &        2 &   50.00 &    50.00 \\
&         \texttt{jp} &       4000 &  269.04 &        2 &   50.00 &    50.00 \\\hline
Hate &     \texttt{en}  &      14100 &  127.14 &        2 &   68.21 &    31.79 \\
Speech  &     \texttt{ge} &       7838 &  149.92 &        2 &   66.64 &    33.36 \\
&   \texttt{hr} &       7839 &  120.74 &        2 &   77.71 &    22.29 \\

&  \texttt{sl} &       7730 &  140.70 &        2 &   55.90 &    44.10 \\
&       \texttt{ar} &       7839 &  104.76 &        2 &   79.60 &    20.40 \\ \midrule

SLU &         \texttt{en}  &      36249 &   35.22 &       12 &   45.77 &     0.40 \\
&          \texttt{es} &       8216 &   39.64 &       12 &   33.17 &     0.03 \\
&         \texttt{th} &       4659 &   31.23 &       10 &   38.68 &     0.59 \\\hline 
XGLUE  &        \texttt{en}  &     119997 & 2887.94 &       10 &   40.96 &     1.39 \\
&         \texttt{ge} &      27998 & 2615.41 &       10 &   30.11 &     0.58 \\

&         \texttt{es}  &      27996 & 3084.47 &       10 &   31.08 &     2.00 \\
&         \texttt{fr}  &      27997 & 2634.95 &       10 &   43.92 &     0.66 \\
&         \texttt{ru}  &      27998 & 2539.33 &       10 &   60.51 &     0.26 \\
\bottomrule
\end{tabular}
}\caption{Dataset Statistics. For each dataset, we provide the total document count, mean document length, and prevalence for the most and least common labels.}
    \label{tab:data_stats}
\end{table}

The motivation for using these datasets is to analyze cross-lingual transfer from multiple perspectives. First, we aim to assess how the selection of source and target languages impacts the transfer. Second, we investigate the effect of class imbalance by including datasets with varying numbers of classes, from binary to 12-class classification. Third, we examine how varying text lengths in datasets influence the results. Lastly, we wanted to ... In addition to language variety, these specific datasets were chosen not only for their diverse linguistic characteristics but also due to their ready availability, representation in prior work, and the specific linguistic challenges they present. By including the Hate Speech and Reviews datasets, we cover sentiment annotation in two distinct domains. The XGLUE News Classification dataset allows us to assess transfer in more formal language settings, while the SLU datasets are used for intent classification.

\subsection{Experimental Setup}
We utilized the XLM-R model \citep{conneau-etal-2020-unsupervised} as the $\theta$ LLM for cross-lingual transfer experiments. We fine-tuned XLM-R in two ways. First, following \citep{ranasinghe2020multilingual,pelicon2021investigating}, we added an extra \textit{classification-head} to XLM-R and fine-tuned all parameters of the model (\emph{289M}). We refer to this strategy as \texttt{full-tune}. Second, we froze the XLM-R weights and added an \textit{adapter-head} \citep{pfeiffer2020mad}. We fine-tuned only the adapter head for a particular task (\emph{1.5M} parameters). We assessed the performance of our models using the macro-averaged $F_1$-score. 

Next, we explain the experimental setups for each research question.

\subsubsection{Adapters vs. Full-model Tuning}
To compare the fine-tuning of all parameters with adapter-only fine-tuning in a cross-lingual setting, we tested three cross-lingual training regimes: \textit{ZS}, \textit{IT}, and \textit{CLV}. This comparison aims to provide insights into the efficacy of different cross-lingual transfer approaches and the contribution of adapters.

\subsubsection{Catastrophic Forgetting in Single Cross-lingual Transfer}
In this experiment, we assessed the effects of forgetting for each cross-lingual transfer strategy (\textit{IT}, \textit{CLV}). For each problem, we first measured the initial performance on the English dataset after fine-tuning on the English training data. Then, for each cross-lingual strategy, we applied the transfer and tested the resulting LLM on the English test data to measure forgetting. The amount of forgetting was expressed as the difference in performance between the final cross-lingually trained models and the initial monolingual English models.

\subsubsection{Catastrophic Forgetting in Multiple Cross-lingual Transfers}
\label{KemkerMetric}
While the previous experiment measured forgetting after transfer to a single language, this experiment assessed forgetting after several steps of cross-lingual transfer, each to a different language. For both full-tuning and adapters, we first trained the model on English, then sorted the target languages based on geographical latitude, reflecting language similarity. For the \texttt{HateSpeech} dataset, we first transferred from \emph{en} to \emph{ge}, then to \emph{sl}, next to \emph{hr}, and finally to \emph{ar}. For the Reviews dataset, the transfer sequence was \emph{en} to \emph{ge}, then to \emph{fr}, and finally to \emph{jp}. For the \texttt{SLU} dataset, the sequence was \emph{en} to \emph{es}, and then to \emph{th}. For the XGLUE dataset, the chain was \emph{en}, \emph{de}, \emph{es}, \emph{fr}, \emph{ru}. After each transfer step, we saved the resulting model and assessed catastrophic forgetting for the previously seen languages. 

We measured performance retention (inverse of forgetting) using the metrics proposed by \citet{kemker2018measuring}. \mrs{Example of the setup for assessing catastrophic forgetting across datasets is shown in Figure \ref{fig:mult_example}.}

\Figure[!t]()[width=0.95\textwidth]{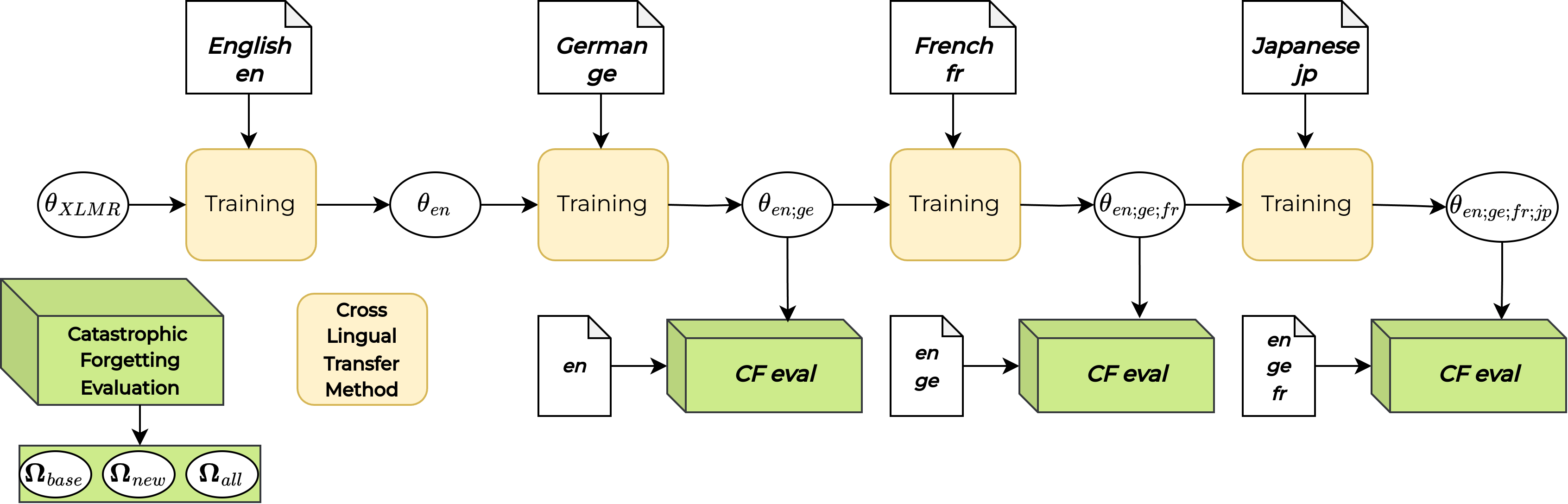}
   {\mrs{Example of the multitransfer approach applied to the Reviews datasets. The cross-lingual method (highlighted in yellow) is first trained or adapted on English and then sequentially transferred to a chain of other languages. After each transfer step, catastrophic forgetting is measured using the Kemker metrics.} \label{fig:mult_example}}

\begin{equation}
    \Omega_{base} = \frac{1}{T-1}\sum^{T}_{i=2}\frac{\alpha_{base,i}}{\alpha_{ideal}}
\end{equation}
\begin{equation}
    \Omega_{new} = \frac{1}{T-1}\sum^{T}_{i=2}\alpha_{new,i}
\end{equation}
\begin{equation}
    \Omega_{all} = \frac{1}{T-1}\sum^{T}_{i=2}\frac{\alpha_{all,i}}{\alpha_{ideal}}
\end{equation}

\noindent 
Here, $\alpha_{ideal}$ is the offline performance, or in our case, the monolingual performance in English $\theta_{ZS}^{\texttt{English}}$. Next, $\alpha_{new, i}$ is the performance of the model on the $i^{th}$ target language (\texttt{tgt}), and $\alpha_{base, i}$ is the model's retention on the \texttt{src} language after the $i^{th}$ out of $T$ sessions. Finally, $\alpha_{all, i}$ is the performance of the model on all previously seen languages at the $i^{th}$ episode.

\subsection{Hyperparameters}
To set the hyperparameters, we utilized the AdamW optimizer \citep{loshchilov2018fixing} for training with an Adam-epsilon of $1e-8$. The \textit{batch-size} was set to \emph{32}, and we employed \textit{early stopping} with a \emph{patience} of 3 steps and a \emph{tolerance} of $0.01$ on the validation loss. The \textit{learning rate} was initialized at $2\cdot10^{-5}$ with \textit{linear scheduling}, warming up for the first $10\%$ of the data, and a \textit{weight decay} of $0.01$. We used predefined seeds $\{1234,1903,42\}$ for reproducibility. \mrs{These seeds were arbitrarily chosen but are commonly used in the research community to facilitate replication of results and assesment of the true mean \cite{reimers-gurevych-2017-reporting}.} 

We developed our experiments using PyTorch Lightning\footnote{\url{https://lightning.ai/docs/pytorch/latest/}} and leveraged the HuggingFace\footnote{\url{https://huggingface.co/}} repository for model implementations. All experiments were conducted on an AMD EPYC 7742 64-Core Processor (utilizing up to 4 cores) with up to two Nvidia A100 GPUs.

\begin{table*}[ht]
\centering
\resizebox{\textwidth}{!}{\begin{tabular}{c|c||cccc|c||ccc|c||cc|c|cccc|c}
\multicolumn{2}{c}{} & \multicolumn{5}{c}{\texttt{HateSpeech}} & \multicolumn{4}{c}{\texttt{Reviews*}} %& \multicolumn{4}{c}{\texttt{Reviews-Music}} & \multicolumn{4}{c}{\texttt{Reviews-Books}} & \multicolumn{3}{c}
&
{\texttt{SLU}} & \multicolumn{5}{c}{\texttt{XGLUE}}  \\ 
fine-t. & mode & \texttt{ge} & \texttt{sl} & \texttt{hr} & \texttt{ar} & $\mu$ &\texttt{ge} & \texttt{fr} & \texttt{jp} & $\mu$  & \texttt{es} & \texttt{th} & $\mu$  & \texttt{ge}  & \texttt{es} & \texttt{fr} & \texttt{ru} & $\mu$ \\ \cline{1-19}

{} &\textit{ZS} & 59.69 & 58.06 & 62.50 & 55.66 & 58.97 &  89.81 & 88.82 & 89.01 & 89.21 & 78.19 &  52.31 & 65.25 & 72.47 &  79.04  & 66.50 & 64.25 & 70.56  \\

{full-tune} &\textit{\textit{IT}}  & \cellcolor{green!25} \textbf{75.36} & \cellcolor{green!25} \textbf{ 76.19} & \cellcolor{green!25} \textbf{ 67.77} & \cellcolor{green!25} \textbf{78.62} & \cellcolor{green!25} \textbf{74.48}  & \cellcolor{green!25} \textbf{91.34} & \cellcolor{green!25} \textbf{90.53} & \cellcolor{green!25} \textbf{90.35} & \cellcolor{green!25} \textbf{90.74} & \cellcolor{green!25} \textbf{89.22} & \cellcolor{green!25} \textbf{95.94} & \cellcolor{green!25} \textbf{92.58} & \cellcolor{green!25} \textbf{78.42} & \cellcolor{green!25} \textbf{ 84.71} & \cellcolor{green!25} \textbf{70.82} & \cellcolor{green!25} \textbf{80.77} & \cellcolor{green!25} \textbf{78.67} \\

{} & \textit{\textit{CLV}}  & 69.71 & 71.84 & 67.27 & 68.63 & 69.36 & 91.0 & 89.49 & 88.44 & 89.64 &  80.68 &  62.19 &  71.43 & 73.43 & 80.90 & 66.79 &  66.04 & 71.78   \\
\hline

{} &\textit{ZS} & 65.04 & 62.38 & \cellcolor{green!25} \textbf{64.08 } & 65.85 & 64.33 & 71.07 & 69.86 & 67.53 & 69.49 &  73.44 & 63.52 & 68.48 & 70.50 &  80.30 & 65.73 & 63.97 & 70.13  \\

\texttt{adapter} & \textit{\textit{IT}}  & \cellcolor{green!25}\textbf{73.37} &\cellcolor{green!25}\textbf{ 73.13} & 56.29 &\cellcolor{green!25} \cellcolor{green!25}\textbf{77.29} & \cellcolor{green!25} \textbf{70.02} & \cellcolor{green!25}\textbf{87.6 }& \cellcolor{green!25} \textbf{86.45} & \cellcolor{green!25} \textbf{85.81 }& \cellcolor{green!25} \textbf{86.62} &  \cellcolor{green!25} \textbf{84.56} & \cellcolor{green!25} \textbf{93.01} & \cellcolor{green!25} \textbf{88.79}& \cellcolor{green!25} \textbf{77.05 }& \cellcolor{green!25} \textbf{83.90 } & \cellcolor{green!25} \textbf{ 69.09 }& \cellcolor{green!25} \textbf{75.53} & \cellcolor{green!25} \textbf{76.39} \\

{} & \textit{\textit{CLV}}  & 67.73 & 68.49 & 62.84 & 68.24 & 66.83 & 87.21 & 82.7 & 86.08 & 85.33 & 74.01 & 67.53 & 70.77 & 72.00 &79.88 & 65.85 &  65.59 & 70.83  \\ \hline

\end{tabular}}
\caption{Results from two fine-tuning methods across three cross-lingual strategies: zero-shot transfer (ZS), intermediate training (\textit{IT}), and cross-validation transfer (\textit{CLV}). English is the source language. The \textit{fine-tuning} column specifies the method (full-tune or adapter), while the $\mu$ column shows the average $F_1$ score across languages. The Reviews dataset aggregates results from three tasks. }
\label{tab:cv_table_all_data}
\end{table*}

\Figure[!t]()[width=0.95\textwidth]{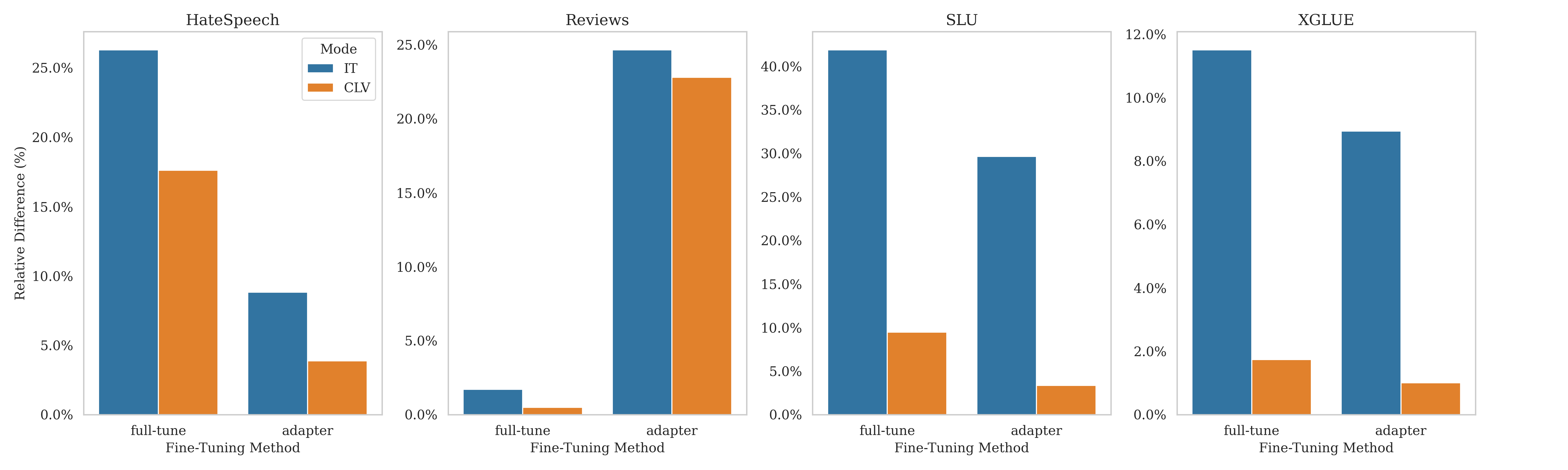}
   {\mrs{Mean relative performance improvements (\%) of full-tuning and adapter-based methods using Intermediate Training (IT) and Cross-Validation Transfer (CLV) 
   strategies compared to the Zero-Shot (ZS) baseline across HateSpeech, Reviews, SLU, and XGLUE datasets. Error bars represent the standard deviation across languages.}\label{fig:relative_diffs}}

\Figure[!t]()[width=0.95\textwidth]{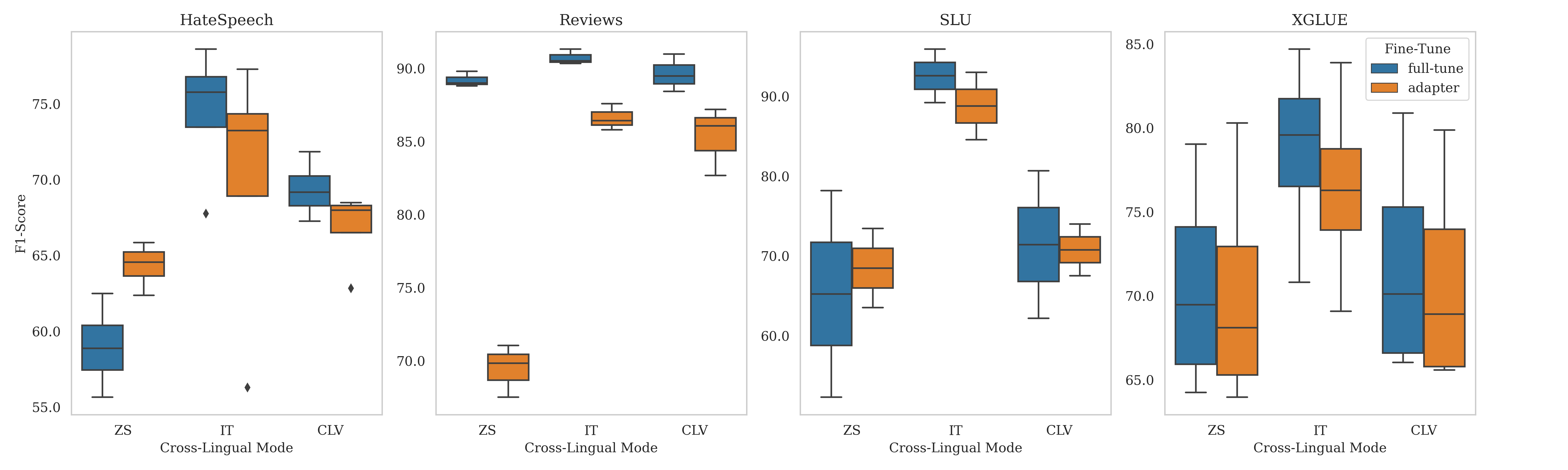}
   {\mrs{Box plots illustrating the performance of full-tuning and adapter-based methods under Zero-Shot (ZS), Intermediate Training (IT), and Cross-Validation Transfer 
   (CLV) conditions for each dataset. Error bars indicate the standard deviation across languages.}\label{fig:boxplots}}

\begin{table*}[ht]
\centering
\resizebox{\textwidth}{!}{
\begin{tabular}{c|c||c|cc||c|cc||c|cc||c|cc||c|cc||c|cc|}
   {fine-t.} &  mode  & \emph{base} & HateSpeech & $\Delta$HateS. & \emph{base} & \texttt{DVD} & $\Delta$DVD  & \emph{base} & \texttt{Music} & $\Delta$Music & base & \texttt{Books} & $\Delta$Books & base & \texttt{SLU} & $\Delta$SLU   & base & \texttt{XGLUE} & $\Delta$XGLUE   \\
       \toprule
  full-tune & \textit{CLV}  & 80.84 & $78.65_{0.004}$ & \cellcolor{green!25} +2.19 &  90.75  & $90.68_{0.004}$ & \cellcolor{red!25}  -0.06 & 91.40  & $90.75_{0.013}$ &\cellcolor{red!25}  -0.65 & 91.75  & $91.56_{0.001}$ & \cellcolor{red!25}  -0.183  & 97.67  & 96.29$_{0.017}$ &  \cellcolor{red!25} -1.38 & 79.15 & 80.21$_{0.013}$  &\cellcolor{green!25} +1.06  \\ 
  {} & \textit{IT}  & 80.84 & $72.85_{0.002}$ &\cellcolor{red!25}  -3.61 & 90.75  & $91.09_{0.002}$ & \cellcolor{green!25} +0.34  &  91.40 & $ 91.72_{0.003}$ &  +\cellcolor{green!25}0.32 & 91.75  & $91.56_{0.006}$ & \cellcolor{red!25}  -0.190 &  97.67  & 92.16$_{0.082}$ & \cellcolor{red!25} -5.51 & 79.15 & 73.32$_{0.020}$ & \cellcolor{red!25} -5.83 \\
  \midrule

  adapter   &  \textit{CLV}  &  76.50  & $75.50_{0.003}$ & \cellcolor{red!25} -1.00 & 64.88 & $58.48_{0.001}$ & \cellcolor{red!25} -6.40 & 64.07 &$60.11_{0.001}$ & \cellcolor{red!25} -3.95  & 88.15 & $60.96_{0.001}$ & \cellcolor{red!25} -27.18 & 93.01 & 85.08$_{0.003}$ & \cellcolor{red!25} -7.93 & 69.37  & 73.93$_{0.016}$ & \cellcolor{green!25} +4.56    \\
  
 {}    &  \textit{IT}  &  76.50 & $60.91_{0.005}$ & \cellcolor{red!25} -15.59 & 64.88 & $59.47_{0.002}$ &\cellcolor{red!25}  -5.41 & 64.07 & $59.72_{0.003}$ & \cellcolor{red!25} -4.35 & 88.15  & $61.37_{0.001}$ & \cellcolor{red!25} -26.78  & 93.01 & 93.38$_{0.016}$ & \cellcolor{green!25} +0.38 & 69.37 & 64.61$_{0.047}$ & \cellcolor{red!25} -4.76  \\ \hline
\end{tabular}
}
\caption{Performance after single cross-lingual transfer on various datasets. Each dataset shows the average macro $F_1$-score, standard deviation, and difference ($\Delta$) from the monolingual English model. Green $\Delta$ values indicate improved performance, while red values indicate a decrease. The \emph{base} column displays the performance of the monolingual English model with the selected fine-tuning for each dataset.}
\label{tab:catastrophic_table}
\end{table*}

\begin{table*}[ht]
    \centering
 \resizebox{0.8\textwidth}{!}{ 
   \begin{tabular}{c|c|c||c c||c c c||cccc}
  &\texttt{tgt} &  \multicolumn{1}{c||}{ge} & \multicolumn{2}{c||}{\texttt{sl}} & \multicolumn{3}{c||}{\texttt{hr}}& \multicolumn{4}{c}{\texttt{ar}} \\ \hline   

  fine-tuning &forgetting in &  \texttt{en} & \texttt{en} & \texttt{ge} & \texttt{en} & \texttt{ge} & \texttt{sl} & \texttt{en} & \texttt{ge} & \texttt{sl} & \texttt{hr} \\ \hline   
  full-tune   & \textit{\textit{IT} mode} & 76.01 & 76.59 & 73.59 & 64.63 & 61.38 & 57.32 & 72.00 & 71.88 & 69.50 & 71.94   \\     
  &      \textit{\textit{CLV} mode} & \cellcolor{green!25} \textbf{79.18} & \cellcolor{green!25} \textbf{76.82} & 71.16 & \cellcolor{green!25} \textbf{67.75} & 58.82 & 56.63 & \cellcolor{green!25} \textbf{ 75.65} & 71.32 & 69.12 & 72.26   \\
\hline
  adapter &   \textit{\textit{IT} mode} & 73.22 & 73.55 & 72.99 & 59.55 & 58.96 & 53.0  &  70.24 & 71.57 & 69.23  & 71.11  \\
    &    \textit{\textit{CLV} mode} & \cellcolor{green!25} \textbf{76.47} & \cellcolor{green!25} \textbf{75.46} &74.22  & \cellcolor{green!25} \textbf{71.3} & 58.77 & 56.63   & \cellcolor{green!25} \textbf{76.34} & 71.21 & 69.13  & 72.27 \\
\hline
    \end{tabular}}
    %}
    \caption{Performance on the HateSpeech dataset for all previous languages after each cross-lingual transfer episode (macro $F_1$-score on the source language(s), to assess the forgetting). We show results for both cross-lingual transfer strategies, \textit{IT} and \textit{CLV}, for the following transfers: \texttt{en}$\to$\texttt{ge}$\to$\texttt{sl}$\to$\texttt{hr}$\to$\texttt{ar}.
    The highlighted results indicate the best performance achieved when applying a respectful fine-tuning approach and cross-lingual transfer technique on the base language (English, in our case).} 
    
    \label{tab:all_multi_transf_hs}
\end{table*}

\begin{table*}[!ht]
    \centering
 \resizebox{0.8\textwidth}{!}{ 
   \begin{tabular}{c|c|c||c c||c c c||cccc}
  &\texttt{tgt} &  \multicolumn{1}{c||}{ge} & \multicolumn{2}{c||}{\texttt{fr}} & \multicolumn{3}{c||}{\texttt{es}}& \multicolumn{4}{c}{\texttt{ru}} \\ \hline   

  fine-tuning &forgetting in &  \texttt{en} & \texttt{en} & \texttt{ge} & \texttt{en} & \texttt{ge} & \texttt{fr} & \texttt{en} & \texttt{ge} & \texttt{fr} & \texttt{es} \\ \hline   
  full-tune   & \textit{\textit{IT} mode} & 74.10 & 68.19 & 73.63 & 74.57 & 75.99 & 71.34 & \cellcolor{green!25} \textbf{ 70.64 }& 67.56 & 64.51 & 77.96 \\

  &      \textit{\textit{CLV} mode} & \cellcolor{green!25} \textbf{81.89} & \cellcolor{green!25} \textbf{70.42} & 71.07 & \cellcolor{green!25} \textbf{ 77.66} & 75.44 & 71.04 & 69.54 & 66.85 & 61.34 & 74.18 \\

\hline
  adapter &   \textit{\textit{IT} mode} & 66.32 & 56.39 & 71.07 & \cellcolor{green!25} \textbf{67.99} & 76.13 & 69.88 & 66.72 & 70.56 & 66.81 & 79.60 \\

    &    \textit{\textit{CLV} mode} & \cellcolor{green!25} \textbf{73.12} & \cellcolor{green!25} \textbf{ 58.17} & 68.65 & 66.46 & 75.9 & 69.43 & \cellcolor{green!25} \textbf{70.39} & 70.5 & 66.63 & 78.73 \\
\hline
    \end{tabular}}
    %}
    \caption{Macro $F_1$-score performance on the XGLUE dataset for all previous languages after each cross-lingual transfer episode (\texttt{en}$\to$\texttt{ge}$\to$ \texttt{fr}$\to$\texttt{es}$\to$ \texttt{ru}).} 
    
    \label{tab:all_multi_transf2_xglue}
\end{table*}
\section{Results}
\label{sec:res}
This section reports the average results of the experimental setups from Section \ref{sec:exp} over three runs. The results are provided in Tables \ref{tab:cv_table_all_data} - \ref{tab:clv_diffs}.

\subsection{Adapters and Full-model Tuning}
We present the results of full fine-tuning and adapters, each using ZS, \textit{IT}, and \textit{CLV} cross-lingual transfer in Table \ref{tab:cv_table_all_data}. Results are reported for all included languages and averaged across all languages.

On average, the \textit{full-tune} learning strategy outperformed the \textit{adapter} training across \textit{ZS} by $8.5\%$, \textit{IT} by $3.81\%$, and \textit{CLV} by $2.85\%$.

In the \textit{full-tune} learning setup, both full-shot scenarios (\textit{CLV} and \textit{IT}) showed improved performance over \textit{ZS}, which served as a strong baseline. This improvement is expected, as the full-shot strategies leverage language-dependent features more effectively during fine-tuning. A similar pattern was observed for \textit{adapter} fine-tuning, except in specific cases: Croatian (\texttt{hr}) on the \textit{HateSpeech} dataset, and Spanish (\texttt{es}) on both the SLU and XGLUE datasets, where zero-shot transfers yielded relatively high scores, and subsequent adapter fine-tuning diminished these results.

On average, the \textit{IT} cross-lingual strategy outperformed \textit{CLV} in almost all experiments, with a $4.46\%$ improvement across the \textit{HateSpeech} datasets and a $1.29\%$ improvement across the \textit{Reviews} datasets. For multilabel datasets, the performance gap was more pronounced, favoring \textit{IT} by $23.82\%$ in \textit{SLU} and $7.18\%$ in \textit{XGLUE}.

In the case of multilabel datasets, \textit{IT} exceeded \textit{CLV} by $19.59\%$ on the \textit{SLU} dataset. This significant difference can be attributed to the variance in data sizes among these datasets. For the \textit{XGLUE} dataset, the performance margin was narrower at $6.22\%$, likely due to the consistent style of news writing and information dissemination across languages \citep{van2010journalism}. Additionally, the data originated from the same period and provider, MSN \citep{liang-etal-2020-xglue}.

For adapters, there were three exceptions where \textit{CLV} outperformed \textit{IT}: \textit{German} and \textit{Japanese} in the \texttt{DVD} category and \textit{Japanese} in the \texttt{Music} category.

In conclusion, if there are no constraints on time and space, or if time is constrained but space is not, we recommend using \textit{full-tune} fine-tuning with the \textit{IT} cross-lingual strategy. On the other hand, if a lower space footprint is required, \textit{adapter} fine-tuning with the \textit{IT} strategy is preferable.

\subsection{Forgetting in Single Transfer}
The forgetting experiments simulate scenarios where a single model is used for multiple languages. While this setup is realistic for full fine-tuning (\textit{full-tune}), it is less realistic for adapters due to their low memory requirements, as storing multiple sets of adapter weights is a common practice. Nonetheless, we measured forgetting for both fine-tuning approaches and reported the results in Table \ref{tab:catastrophic_table}, showing forgetting with a single cross-lingual transfer for different datasets.

\mrs{In Figure~\ref{fig:relative_diffs}, we show the relative differences between the selected cross-lingual (CL) transfer methodologies and zero-shot transfer across different tuning modes for each dataset. In Figure~\ref{fig:boxplots}, we present the distribution of scores for each dataset as evaluated by the cross-lingual method and transfer paradigm.}

For the \textit{Reviews} dataset, we observed an increase in performance for \textit{full-tune IT} transfer in the \texttt{DVD} and \texttt{Music} categories, while forgetting occurred in the remaining combinations. Forgetting was minimal for the \textit{full-tune} approach (around $1\%$), but ranged from moderate (for \texttt{DVD} and \texttt{Music}) to significant (for \texttt{Books}) with \textit{adapter} fine-tuning. The higher forgetting rates in \texttt{Books} stemmed from a stronger initial performance baseline, whereas \texttt{DVD} and \texttt{Music} showed more moderate forgetting due to lower initial baselines.

For the \textit{HateSpeech} dataset, cross-lingual validation generally decreased performance, except for \textit{full-tune} transfer with the \textit{CLV} method, which achieved a $2.19\%$ gain.

For the \textit{SLU} dataset, both fine-tuning modes yielded strong monolingual models. However, performance degraded for both \textit{IT} and \textit{CLV} approaches due to discrepancies in dataset sizes across languages. Using adapters, the \textit{CLV} paradigm improved performance by $0.38\%$. For the \textit{XGLUE} dataset, \textit{CLV} yielded a $1.06\%$ improvement with \textit{full-tune} and a $4.56\%$ improvement with adapters, likely due to the structured nature and consistent writing style of the dataset.

%\subsection{Results of Forgetting in Multiple Cross-lingual Transfers}
\subsection{Forgetting in Multiple Transfers}

Table \ref{tab:multi_clp_assesment} shows performance retention (inverse of forgetting) with the metrics of \citet{kemker2018measuring}, defined in Section \ref{KemkerMetric}. For both the binary and multilabel classification datasets, we see that with the \textit{\textit{CLV}} transfer strategy, measured with the $\Omega_{base}$, most of the performance in the base language, English is preserved (and improved for the \textit{Reviews} and \textit{XGLUE} datasets). A notable difference in the impact of the tuning strategy is evident in the \textit{SLU} dataset, attributed to language differences and discrepancies in data size.

\begin{table}[htb]
    \centering
\resizebox{0.70\linewidth}{!}{\begin{tabular}{l|l|rrr}
\toprule
fine-tuning & mode &   $\Omega_{base}$ &    $\Omega_{new}$ &    $\Omega_{all}$ \\
\hline
\multicolumn{5}{c}{\texttt{HateSpeech}} \\
\hline
full-tune &     \textit{IT} & 94.56 & \cellcolor{green!25} \textbf{71.59} & \cellcolor{green!25} \textbf{92.86} \\

 & \textit{CLV} & \cellcolor{green!25} \textbf{97.89} & 69.55 & 91.88 \\  
 \hline
 adapter &    \textit{IT} & 68.86 & \cellcolor{green!25} \textbf{67.69} & \cellcolor{green!25} \textbf{65.80} \\
 &   \textit{CLV} & \cellcolor{green!25} \textbf{72.32} & 67.01 & 63.70 \\
\hline
\multicolumn{5}{c}{\texttt{Reviews-Dvd}} \\
\hline
full-tune &    \textit{IT} & \cellcolor{green!25} \textbf{100$^{*}$} & \cellcolor{green!25} \textbf{90.26 }& 66.36 \\
&   \textit{CLV} & \cellcolor{green!25} \textbf{100$^{*}$}  & 89.95 & \cellcolor{green!25} \textbf{65.25} \\
\hline
adapter &    \textit{IT} & 92.56 & 68.40 & 73.27 \\ 
&   \textit{CLV} & \cellcolor{green!25} \textbf{93.07} & \cellcolor{green!25} \textbf{68.56 }& \cellcolor{green!25} \textbf{73.59} \\
 
\toprule
\multicolumn{5}{c}{\texttt{Reviews-Music}} \\
\toprule
full-tune  &    \textit{IT} & 78.95 & 81.67 & 54.24 \\ 
&   \textit{CLV} & \cellcolor{green!25} \textbf{100$^{*}$}  & \cellcolor{green!25} \textbf{91.00} & \cellcolor{green!25} \textbf{66.53} \\
\hline
adapter&    \textit{IT} & \cellcolor{green!25} \textbf{100$^{*}$}  & \cellcolor{green!25} \textbf{78.61} & \cellcolor{green!25} \textbf{91.97}   \\
&   \textit{CLV} & 94.56 & 69.43 & 75.43 \\
\toprule
\multicolumn{5}{c}{\texttt{Reviews-Books}} \\
\hline
full-tune &    \textit{IT} & \cellcolor{green!25} \textbf{100$^{*}$}   & \cellcolor{green!25} \textbf{91.54} & \cellcolor{green!25} \textbf{66.80 }\\ 
&   \textit{CLV} & \cellcolor{green!25} \textbf{100$^{*}$}  & 91.13 & 66.67 \\
\hline
adapter &    \textit{IT} & \cellcolor{green!25} \textbf{91.13} & \cellcolor{green!25} \textbf{79.83} & \cellcolor{green!25} \textbf{67.71} \\
&   \textit{CLV} & 90.70 & 79.17 & 67.23 \\
\hline
\multicolumn{5}{c}{\texttt{SLU}} \\
\hline
full-tune &    \textit{IT} & \cellcolor{green!25} \textbf{32.63} & \cellcolor{green!25} \textbf{61.25} & \cellcolor{green!25} \textbf{46.71} \\
&   \textit{CLV} & 29.89 & 57.09 & 41.48\\
\hline
adapter &    \textit{IT} &  \textbf{32.81} & \cellcolor{green!25} \textbf{59.09} & \cellcolor{green!25} \textbf{46.53}\\
&   \textit{CLV} & 30.21 & 55.79 & 41.93 \\
\hline
\multicolumn{5}{c}{\texttt{XGLUE}} \\
\hline
full-tune &    \textit{IT} & \cellcolor{green!25} \textbf{100$^{*}$}   & \cellcolor{green!25} \textbf{100$^{*}$} & \cellcolor{green!25} \textbf{92.53}\\ 
&   \textit{CLV} & \cellcolor{green!25} \textbf{100$^{*}$}  & 99.14 & 92.10 \\
\hline
adapter &    \textit{IT} & \cellcolor{green!25} \textbf{100$^{*}$} & \cellcolor{green!25} \textbf{100$^{*}$} & \cellcolor{green!25} \textbf{100$^{*}$} \\
&   \textit{CLV} &  \textbf{100$^{*}$} & 97.75 & \cellcolor{green!25} \textbf{100$^{*}$}  \\
\hline
\end{tabular}}

    %\caption{The performance retention using \textit{full-tune} vs. \textit{adapter} cross-lingual transfer with \textit{IT} and \textit{CLV} strategies across multiple transfers is presented. We measured the results described in Section \ref{KemkerMetric}, expressed them as percentages, and capped them at 100*. The best performance for each dataset, considering tuning strategy and cross-lingual mode, is indicated in bold.}
    \caption{We compared \textit{full-tune} vs. \textit{adapter} cross-lingual transfer using \textit{IT} and \textit{CLV} strategies across multiple transfers. Results are expressed as percentages capped at 100*. The best performance for each dataset, considering tuning strategy and cross-lingual mode, is highlighted in bold.
    }
    \label{tab:multi_clp_assesment}
\end{table}

Due to its unique status as a lingua franca, performance retention relative to English might be a special case. Namely, across all cross-lingual transfers (measured with $\Omega_{all}$), as well as in acquiring new knowledge (measured with $\Omega_{new}$), the \textit{IT} strategy is usually better.

\begin{table}[t!]
    \centering
    \resizebox{0.9\columnwidth}{!}{
   \begin{tabular}{c|c|c||c c||c c c}
 & \texttt{tgt} &  \multicolumn{1}{c||}{ge} & \multicolumn{2}{c||}{\texttt{fr}} & \multicolumn{3}{c}{\texttt{jp}} \\ \hline   
 {} &  forget. in &  \texttt{en} & \texttt{en} & \texttt{ge} &\texttt{en} & \texttt{ge} & \texttt{fr} \\ \cline{2-8}
 {fine-tuning} &  mode & \multicolumn{6}{c}{}  \\ 
 \midrule
 \multicolumn{8}{c}{\texttt{Reviews-Books}} \\ \hline \hline
full-tune & \textit{\textit{IT}} & 90.98 & \cellcolor{green!25} \textbf{92.67} & 92.17 & \cellcolor{green!25} \textbf{92.52} & 91.77 & 91.51 \\

&   \textit{\textit{CLV}} & \cellcolor{green!25} \textbf{91.50} & 92.05 & 92.05 & 92.50 & 92.10 & 91.35 \\
\hline

adapter & \textit{\textit{IT}} & \cellcolor{green!25} \textbf{61.41} & \cellcolor{green!25} \textbf{ 90.04} & 89.79 & \cellcolor{green!25} \textbf{89.97} & 89.48 & 89.19 \\

&   \textit{\textit{CLV}} & 60.96 & 89.27 & 89.45 & 89.12 & 88.94 & 88.57 \\

\midrule
 \multicolumn{8}{c}{\texttt{Reviews-DVD}} \\ \hline \hline

full-tune &  \textit{\textit{IT}} & \cellcolor{green!25} \textbf{91.29} & \cellcolor{green!25} \textbf{90.52} & 91.37 & 90.27 & 89.71 & 90.02 \\
&   \textit{\textit{CLV}} & 90.25 & 89.34 & 91.27 & \cellcolor{green!25} \textbf{90.45 }& 91.37 & 90.31 \\ \hline

adapter & \textit{\textit{IT}}& 59.68 & 86.64 & 60.24 & 86.99 & 60.41 & 87.31 \\

&   \textit{\textit{CLV}} & \cellcolor{green!25} \textbf{ 60.11} & \cellcolor{green!25} \textbf{87.12} & 60.52 & \cellcolor{green!25} \textbf{87.57} & 60.79 & 87.49 \\
\midrule
 \multicolumn{8}{c}{\texttt{Reviews-Music}} \\ \hline \hline

full-tune & \textit{\textit{IT}} & \cellcolor{green!25} \textbf{91.80 }& \cellcolor{green!25} \textbf{91.95} & 62.34 & 91.37 & 61.86 & 90.91 \\
&   \textit{\textit{CLV}} & 91.47 & 91.62 & 91.62 & \cellcolor{green!25} \textbf{91.62} & 91.15 & 90.85 \\
\hline

adapter & \textit{\textit{IT}}& 59.76 & 88.70 & 88.20 & 88.66 & 88.74 & 88.45 \\

&   \textit{\textit{CLV}} & \cellcolor{green!25} \textbf{60.11} & \cellcolor{green!25} \textbf{88.72} & 60.82 & \cellcolor{green!25} \textbf{89.11} & 61.03 & 88.65 \\

\hline
    \end{tabular}
    }
    \caption{Macro $F_1$-score performance on the Reviews datasets for all previous languages after each cross-lingual transfer episode 
 (\texttt{en}$\to$\texttt{ge}$\to$\texttt{fr}$\to$\texttt{ru}) .}
    %We show results for both cross-lingual transfer strategies, \textit{IT} and \textit{CLV}, making the transfer: English $\to$ German $\to$ French $\to$ Japanese. The highlighted results indicate the best performance achieved when applying a respectful fine-tuning approach and cross-lingual transfer technique on the base language (English, in our case). }
    
    \label{tab:all_multi_transf_all}
\end{table}

In Table \ref{tab:all_multi_transf_hs}, we present the evaluation results for multiple transfers in the \textit{HateSpeech} dataset. Measuring forgetting relative to English as the source language across all transfer steps, the \textit{CLV} cross-lingual transfer strategy consistently outperformed \textit{IT}, with an average improvement of $2.54\%$. Conversely, forgetting relative to other languages was lower with the \textit{IT} strategy (except for the Croatian-to-Arabic transfer), with an average improvement of $1.05\%$. Even in longer transfer chains, forgetting relative to English remained the lowest. This could be attributed to English's strong global cultural influence, which is also reflected in patterns of hate speech. Using \textit{adapter} fine-tuning yielded results consistent with the \textit{full-tune} approach for the \textit{HateSpeech} dataset.

For the \textit{Reviews} datasets (Table \ref{tab:all_multi_transf_all}), we observed enhanced retention when employing the \textit{IT} strategy with \textit{full-tune} fine-tuning for \texttt{DVD} and \texttt{Music}. The \textit{IT} strategy was also beneficial for \textit{adapters}, helping to preserve a substantial amount of information for \texttt{Music} and \texttt{Books}. Both cross-lingual paradigms in the two fine-tuning scenarios demonstrated performance improvements over the base model.

We observed a similar retention trend for the XGLUE multilabel dataset (Table \ref{tab:all_multi_transf2_xglue}). The \textit{CLV} strategy surpassed the \textit{IT} approach in most cases, except when fine-tuned on the Russian dataset, where both strategies performed comparably. This may be due to the linguistic and script variations between languages \citep{fujinuma-etal-2022-match}.

In the \textit{SLU} scenario (Table \ref{tab:all_multi_transf_slu}), both the adapter and full-tune methods effectively retained information using the \textit{IT} strategy. Notably, a significant difference was observed in the adapter's retained knowledge when adapted to the Thai language compared to Spanish and English, likely due to differences in data quantity.

\begin{table}[htb]
    \centering
    \resizebox{0.7\columnwidth}{!}{
   \begin{tabular}{c|c|c||c c}
 & \texttt{tgt} &  \multicolumn{1}{c||}{es} & \multicolumn{2}{c}{\texttt{th}}  \\ \hline   
 {} &  forgetting in &  \texttt{en} & \texttt{en} & \texttt{es} \\ \cline{2-5}
 {fine-tuning} &  mode & \multicolumn{3}{c}{}  \\ 
 \hline
full-tune & \textit{\textit{IT}} & \cellcolor{green!25} \textbf{97.55} & \cellcolor{green!25} \textbf{95.60} & 88.90 \\

&   \textit{\textit{CLV}} & 86.30 & 87.58 & 79.08 \\ \hline

adapter & \textit{\textit{IT}} & \cellcolor{green!25} \textbf{94.56} & \cellcolor{green!25} \textbf{91.54} & 83.56 \\

&   \textit{\textit{CLV}} & 85.35 & 84.28 & 75.72  \\

    \end{tabular}
    }
    \caption{Macro $F_1$-score performance on the SLU datasets for all previous languages after each cross-lingual transfer episode (\texttt{en} $\to$ \texttt{es} $\to$ \texttt{th}). }
    
    \label{tab:all_multi_transf_slu}
\end{table}

\subsection{Validation Set Structure in \textit{CLV}}

The composition of the validation set plays an important role in the \textit{CLV} cross-lingual transfer strategy. Recall that we can either use only the validation set from the target language, $L_{\texttt{tgt}}^{\texttt{valid}}$ (valid approach), or merge the target training and validation sets to form a combined validation set: $L_{\texttt{tgt}}^{\texttt{merged}} = L_{\texttt{tgt}}^{\texttt{train}} \cup L_{\texttt{tgt}}^{\texttt{valid}}$ (valid+train approach). Table \ref{tab:clv_diffs} compares these two approaches for both full-tune and adapter fine-tuning.

For the full-tune approach, the differences between the two validation strategies are small. However, for adapters, the differences are significant, particularly in binary classification datasets like \textit{Reviews} and \textit{HateSpeech}. On average, the valid+train approach results in a 7.73\% improvement. This improvement can be attributed to the larger validation set, which allows for more reliable updates to the adapter features. In contrast, the full-tune approach adapts all model weights, making it more robust and less sensitive to the size of the validation set.

For larger multilabel datasets, specifically \textit{XGLUE} and \textit{SLU}, where the training data significantly exceeds the validation data in magnitude and label count, enlarging the validation set offers minimal benefits. Both the \textit{adapter} and \textit{full-tune} approaches show only marginal gains of approximately $0.1\%$ in such scenarios.

\begin{table}[htb]
    \centering
    \resizebox{0.8\columnwidth}{!}{\begin{tabular}{c|cc}
        dataset/fine-tuning & \texttt{full-tune} & \texttt{adapters} \\ \hline 
        \multicolumn{3}{c}{binary dataset} \\
        \hline 
        \texttt{HateSpeech} & -0.03 & 9.18   \\ 
        \texttt{Reviews-DVD} & 0.06  & 8.84 \\ 
        \texttt{Reviews-Music} & 0.30 &  8.91 \\ 
        \texttt{Reviews-Books} & 0.11 & 3.97  \\ 
        \hline
        avg. improvement & 0.11 & 7.73 \\ 
        \hline 
        \multicolumn{3}{c}{multi-label datasets} \\
        \hline
        \texttt{SLU} & 0.02 & 0.02 \\ 
        \texttt{XGLUE} & 0.00 & 0.01 \\ \hline \hline
        avg. improvement & 0.01 & 0.01 \\ \hline
    \end{tabular}}
    %}
    \caption{Differences in performance measured by macro $F_1$-score for the \textit{CLV} cross-lingual transfer regarding the \emph{validation} set size. The results show the average difference in performance (\%) between the  \textit{valid} approach and the \textit{valid+train} one.}
    \label{tab:clv_diffs}
\end{table}

\subsection{Computational Efficiency} % of Cross-lingual Transfer}

%\begin{table}[H]
%    \centering
%    \resizebox{\linewidth}{!}{\begin{tabular}         {c|c c | c c| c c}
%        {} & \multicolumn{2}{c|}{avg. time per epoch} & \multicolumn{2}{c|}{avg. epochs} & \multicolumn{2}{c}{avg. total time} \\      
%        {fine-tuning} & full-tune   &adapter & full-tune   &adapter & full-tune & adapter\\ \hline   
%        ZS & 53.20s & {50.80}s & \cellcolor{green!25} \textbf{5.75}  &  7.25 & 305.91s & 368.32s\\  
%        \textit{IT} & {110.35}s & \cellcolor{green!25} \textbf{{103.51}s} & 4.05 & 6.44 & \cellcolor{green!25} \textbf{537.35s} & \cellcolor{green!25} \textbf{707.75s} \\ 
%        \textit{CLV}(\textit{v+t}) & \cellcolor{green!25} \textbf{ {117.22s}} & {87.19s} & 4.16 & \cellcolor{green!25} \textbf{7.88}& 487.63s  & 687.08s \\ 
%        \textit{CLV}(\textit{v}) & {95.19}s & {60.74}s & 4.23 & 7.77 & 409.32s & 473.77s \\ 
%        \hline
%        \end{tabular}}
%    \caption{Total time per training run, average computational time per epoch (in seconds), and the average number of epochs until convergence for each fine-tuning approach and cross-lingual strategy. The highlighted outcomes indicate the metrics that require the highest computational resources concerning the specified criteria. 
%    }
%    \label{tab:comp_times}
%\end{table}

\begin{table}[htb]
    \centering
    \resizebox{\linewidth}{!}{\begin{tabular}{l|cc|cc|cc}
     & \multicolumn{2}{c|}{avg. epochs} & \multicolumn{2}{c|}{avg. time per epoch} & \multicolumn{2}{c}{avg. total time} \\
    model & adapter & full & adapter & full & adapter & full \\
    \hline
    \textit{ZS} & 7.67 & 5.33 & 266.54s & 542.53s & 2043.45s & 2893.47s \\
    \textit{IT} & 6.96 &  \textbf{5.37} & 71.17s & 124.19s &  \textbf{2538.81s} & \textbf{3559.94s} \\
    \textit{CLV(v+t)} &  \textbf{8.17} & 4.61 &  \textbf{313.48s} & \textbf{ 677.02s} & 2561.16s & 3118.79s \\
    \textit{CLV(v)} & 8.10 & 4.40 & 286.37s & 640.51s & 2318.67s & 2820.36s \\
    \end{tabular}}

    \caption{Number of epochs until convergence, average computational time per epoch (in seconds), and the total time for each combination of fine-tuning and cross-lingual strategy. The highlighted outcomes indicate the metrics that require the highest computational resources concerning the specified criteria. \texttt{NB:} the total time for the \textit{IT} represents the average time needed for source and transfer training.}
    \label{tab:comp_times}
\end{table}

Table \ref{tab:comp_times} presents the computational time for different fine-tuning methods and cross-lingual transfer strategies. We report the average times for training and validation in a single epoch, along with the number of epochs required for convergence. On average, the \texttt{adapter} method required $2.8$ more epochs to converge but $261.42$ seconds less per epoch.

When updating all model parameters (\texttt{full-tune}), the \textit{IT} strategy required approximately $82\%$ less time per epoch compared to the \textit{CLV} strategy with the \textit{valid+train} (\textit{v+t}) approach, primarily due to smaller training and validation sets. However, in terms of total time per run, the \textit{IT} approach (which involves monolingual fine-tuning on the \texttt{source} language followed by tuning on the \texttt{target} language) required an additional $441.24$ seconds compared to \textit{CLV} for full-tune. For adapter-based methods, \textit{IT} tuning added an extra $22.35$ seconds compared to \textit{CLV}.

\Figure[t!](topskip=0pt, botskip=0pt, midskip=0pt)[width=0.9\linewidth] {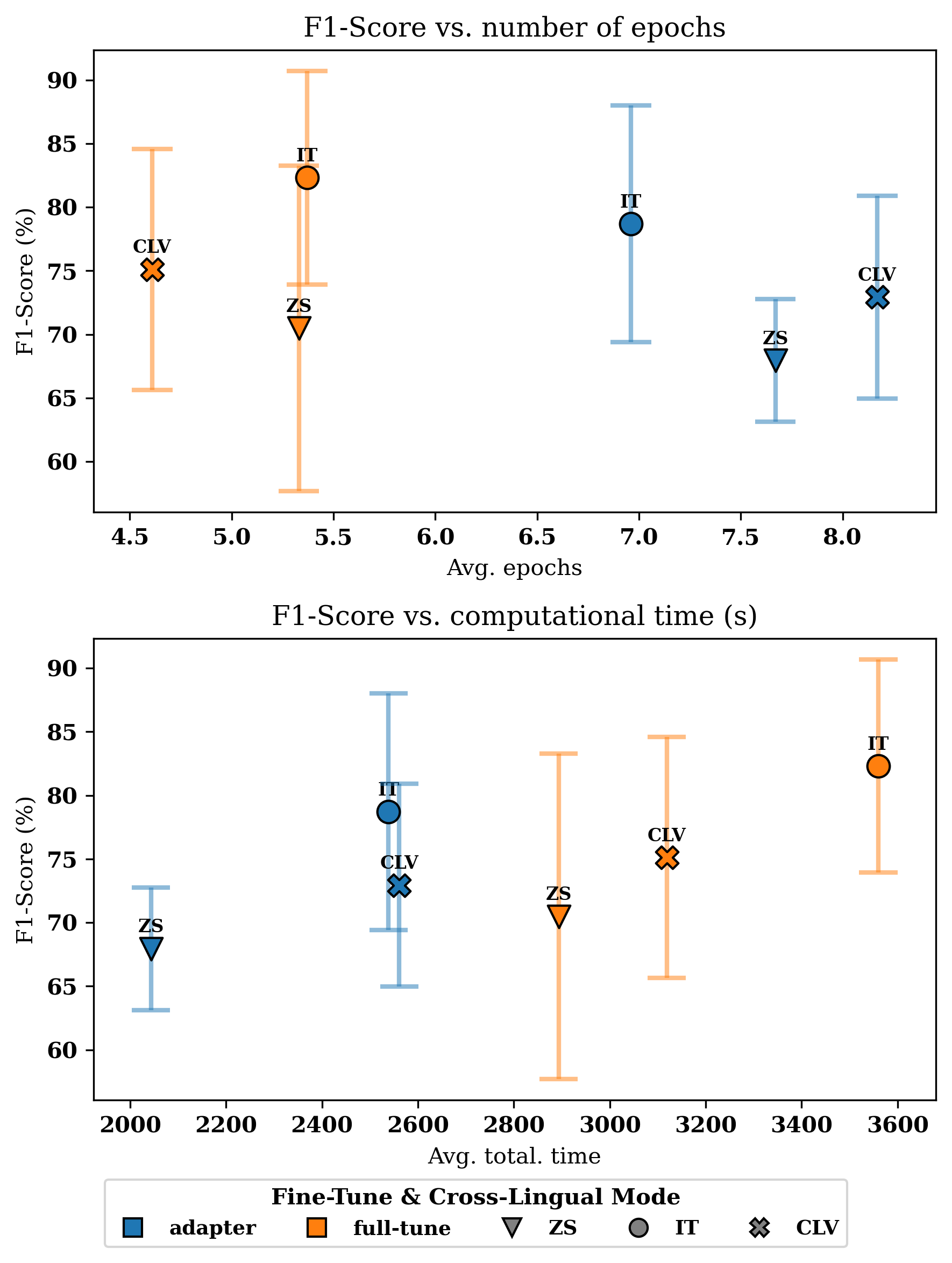} {\mrs{Comparison of the average $F_1$-score per method versus the average number of epochs and total running time.} \label{fig:time}}

\mrs{In Figure~\ref{fig:time}, we show the interaction between the $F_1$-score, average training time, and the number of epochs required. On average, adapter-based methods require more epochs but complete training faster overall, leading to lower performance for cross-lingual transfer tasks. By contrast, fully fine-tuned methods, though more computationally expensive and time-intensive, particularly the \textit{IT} method, achieve better performance. These results highlight the trade-offs between computational efficiency and model performance, which depend on the chosen fine-tuning strategy and task-specific requirements. For scenarios demanding high cross-lingual performance, full-tuning methods, particularly the \textit{IT} variant, may be preferred despite their higher computational cost. On the other hand, adapter-based methods offer a viable alternative under computational or time constraints, albeit at the cost of reduced performance in tasks like cross-lingual transfer. The variability in $F_1$-scores across configurations underscores the importance of considering task characteristics and resource constraints when selecting fine-tuning strategies.}

\section{Conclusion and Further work}
\label{sec:final}
Our empirical study explored various cross-lingual transfer learning strategies in combination with two fine-tuning approaches of LLMs. We based our findings on multiple classification problems, each represented with datasets in several languages. We first investigated the impact of the cross-lingual training strategies and compared the effectiveness of \textit{\textit{CLV}} and \textit{\textit{IT}}. Results show that cross-lingual transfer with intermediate training, which uses languages sequentially, is more effective than \textit{CLV} transfer, which uses target languages directly as a validation set.

The second set of experiments examined how cross-lingual transfer affects forgetting in the source language. We found that forgetting is generally comparable between \textit{IT} and \textit{CLV}. However, in multiple cross-lingual transfers, the \textit{CLV} strategy better mitigates catastrophic forgetting and retains more knowledge from the source language than the \textit{IT} method. %
Nevertheless, possibly due to global cultural presence, the forgetting in English is a special case and lower than other languages. The retention of knowledge in English is better with the \textit{CLV} strategy, while for other languages and across several cross-lingual steps, the \textit{IT} strategy causes less forgetting. We produce language-tuned and multilingual adapter modules
for each task that other researchers can reuse via AdapterHub \citep{pfeiffer2020adapterhub}. Regarding computation, we observed that the adapter and full-tune tuning using the \textit{IT} require more total time and epochs to converge than the \textit{CLV}. 

In future work, expanding the scope of our experiments to include a wider range of languages and problems, specifically incorporating low-resource datasets, would enhance the generality of our findings. We recommend leveraging additional knowledge sources from expansive knowledge banks such as BabelNet \citep{navigli2012babelnet} to further enrich the learning process. Furthermore, we suggest evaluating the occurrence of catastrophic forgetting in graph-based document approaches. We hypothesize that, by leveraging neighborhood sharing, certain local knowledge can be acquired and transferred across different languages.

\section*{Limitations}
The scope of our study is restricted to the family of classification problems. This implies that our findings may not directly apply to other problems. Further, we focus on English as the base source language; the transfer between others, especially similar languages, may differ. Although adapter fusion has proven to be a highly effective method for amalgamating knowledge from multiple learned tasks to solve new problems, we could not explore this approach due to time constraints. Therefore, our study may not have fully captured the potential benefits of adapter fusion in cross-lingual transfer. Additionally, potential biases inherent in dataset sources—such as those between social media and news outlets—might influence cross-lingual transfer performance by introducing variations in formality, vocabulary, and stylistic nuances.

\section*{Ethics Statement}
The authors have used only existing datasets and do not identify any novel elements for ethical considerations. However, even when using the trained models from existing datasets, any user should be aware that the models are not performing without mistakes.

\bibliographystyle{apalike}

\bibliography{custom}

\begin{IEEEbiography}
[{\includegraphics[width=1in,height=1.25in,clip,keepaspectratio]{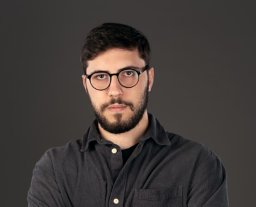}}] Boshko Koloski is a Ph.D. student at the Jožef Stefan International Postgraduate School, specializing in learning from heterogeneous data sources, from texts to graphs. His research lies at the intersection of natural language processing and graph learning, with a particular focus on machine learning in low-resource settings for underrepresented languages. Boshko is especially interested in cross-lingual transfer learning, driven by the process of transferring acquired knowledge.

\end{IEEEbiography}

\begin{IEEEbiography} [{\includegraphics[width=1in,height=1.25in,clip,keepaspectratio]{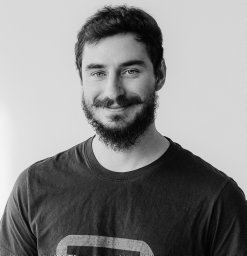}}] Bla\v{z} \v{S}krlj is an assistant professor at the Jo\v{z}ef Stefan
International Postgraduate School. He focuses on representation learning at scale. He obtained his Ph.D. at the Jo\v{z}ef Stefan International Postgraduate School in 2023 under supervision of prof. dr. Nada Lavra\v{c}.

\end{IEEEbiography}
\begin{IEEEbiography} [{\includegraphics[width=1in,height=1.25in,clip,keepaspectratio]{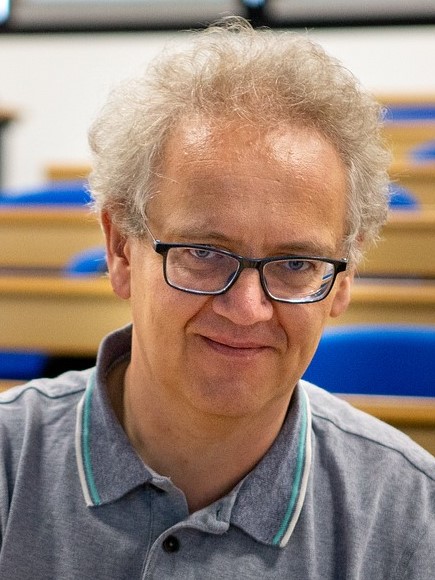}}] Marko Robnik-Šikonja is a Professor of computer science and informatics and Head of Machine Learning and Language Technologies Lab at the Faculty of Computer and Information Science, University of Ljubljana. His research interests include machine learning, data mining, natural language processing, network analytics, and the application of data science techniques. His most notable scientific results are from the areas of feature evaluation, ensemble learning, explainable artificial intelligence, data generation, and natural language analytics. He is the (co)author of over 300 scientific publications that were cited more than 9,000 times, several open-source machine learning packages, large language models, and language resources. He participates in several national and international projects, regularly serves as a program committee member of top artificial intelligence and machine-learning conferences, and is an editorial board member of seven international journals.

\end{IEEEbiography}
\begin{IEEEbiography} [{\includegraphics[width=1in,height=1.25in,clip,keepaspectratio]{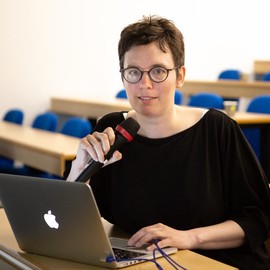}}] 
Senja Pollak is a researcher at the Jožef Stefan Institute and an assistant professor at the Jožef Stefan International Postgraduate School. Her work centers on natural language processing, with research interests spanning text mining, corpus linguistics, and computational creativity. She also teaches courses on language technologies and computational creativity at the Jožef Stefan International Postgraduate School. From 2018 to 2019, she was a research fellow at the Usher Institute, University of Edinburgh. Senja has actively contributed to various conferences, serving as a co-chair for SLSP 2019 and for workshops such as Hackashop on news media content analysis and automated report generation at EACL 2021, BSNLP at EACL 2021, and Digital Humanities and Natural Language Processing at PROPOR 2020. Additionally, she has been a program committee member for ICCC, JTDH, and SYNASC 2021, and a reviewer for LREC, ACL-IJCNLP 2021, among others.

\end{IEEEbiography}

\EOD

\end{document}